\documentclass{article}
\usepackage{ijcai20}

\usepackage{times}
\usepackage{soul}
\usepackage{url}
\usepackage[hidelinks]{hyperref}
\usepackage[utf8]{inputenc}
\usepackage[small]{caption}
\usepackage{graphicx}
\usepackage{amsmath}
\usepackage{amsthm}
\usepackage{booktabs}
\usepackage{algorithm}
\usepackage{algorithmic}
\usepackage{amsfonts,amssymb}
\usepackage{enumitem}
\usepackage{subfigure}
\usepackage{amsmath, amssymb}
\usepackage{makecell}
\usepackage{multirow}

\usepackage[usenames, dvipsnames]{color}
\usepackage{tabularx}
\usepackage[font=small,skip=0pt]{caption}
\usepackage{graphicx}
\usepackage{varwidth}

\newcommand{\blue}[1]{{\color{blue} #1}}

\title{Heterogeneous-Temporal Graph Convolutional Networks: Make the Community Detection Much Better}

\author{
Yaping Zheng$^1$\footnote{Contact Author}\and
Shiyi Chen$^1$\and
Xinni Zhang$^2$\and
Xiaofeng Zhang$^1$\and
Xiaofei YANG$^3$\and
Di Wang$^4$\\
\affiliations
$^1$Harbin Institute of Technology\\
$^2$Jilin University\\
$^3$Peng Cheng Laboratory\\
$^4$Nanyang Technological University, Singapore\\
\emails
zhengyaping.hit@qq.com,
chenshiyi.hit@gmail.com,
zhangxn2116@mails.jlu.edu.cn,
zhangxiaofeng@hit.edu.cn,
xiaofei.hitsz@gmail.com,
wangdi@ntu.edu.sg
}

\begin{document}

\maketitle

\begin{abstract}
Community detection has long been an important yet challenging task to analyze complex networks with a focus on detecting topological structures of graph data. Essentially, real-world graph data contains various features, node and edge types which dynamically vary over time, and this invalidates most existing community detection approaches. 
To cope with these issues, this paper proposes the heterogeneous-temporal graph convolutional networks (HTGCN) to detect communities from hetergeneous and temporal graphs. 
Particularly, we first design a heterogeneous GCN component to acquire feature representations for each heterogeneous graph at each time step. 
Then, a residual compressed aggregation component is proposed to represent ``dynamic'' features for ``varying'' communities, which are then aggregated with ``static'' features extracted from current graph.  
%
Extensive experiments are evaluated on two real-world datasets, i.e., DBLP and IMDB. The promising results demonstrate that the proposed HTGCN is superior to both benchmark and the state-of-the-art approaches, e.g., GCN, GAT, GNN, LGNN, HAN and STAR, with respect to a number of evaluation criteria. 

\end{abstract}

\section{Introduction}
Community detection \cite{fortunato2016communityreview} has long been an important yet challenging task to analyze complex networks \cite{borgatti2009socialnetwork,cavallari2017dblp} with a focus on detecting topological structures of homogeneous graphs with flourishing results \cite{girvan2002community,blei2003lda,newman2004modularity,lee2001NMF,he2018MRF,shao2019clustering}.
However, the real-world attributed graphs are generally heterogeneous, 
and dynamically varying over time, which pose great challenges to most existing community detection approaches. 

Notably, there exist very few related community detection methods  \cite{du2018galaxy,li2019embeddingcommunity,dall2019het-com} to simultaneously address these issues under a unified framework.  However, a vast amount of research effort could be seen in the literature 
with a focus on feature (or attribute) embedding \cite{kipf2017semi-supervised}, heterogeneous graph analysis \cite{peng2019heterogeneousconvolutional} or temporal data prediction \cite{singer2019temporalgraph}.
For feature embedding task, various graph neural network (GNN) based approaches have been proposed \cite{spectral2014joan,grover2016node2vec,kipf2017semi-supervised,hamilton2017inductive,velikovi2018GAT,you2019position-aware} to embed attributes or spatial information of graphs. 
In \cite{spectral2014joan}, a graph Fourier transformation is defined to perform spectral convolutions operated on non-Euclidean graph data. 
GraphSAGE learns an inductive feature representation method \cite{hamilton2017inductive} which could be used to predict class labels of unseen nodes. 
For heterogeneous graph analysis,  
\cite{Wang:2019:HAN} proposes a node-level and a semantic-level attention component to assign weights to heterogeneous nodes contained in different layers. To perform temporal data prediction, STAR \cite{xu2019star} proposes a RNN based approach which learns both temporal and spatial feature representations. 
Among all these approaches, the line graph neural network (LGNN) \cite{chen2019supervised} is the state-of-the-art deep learning based approach, with powerful feature representation ability, for homogeneous community detection. LGNN cast the 
community detection problem to node-wise classification task by the proposed permutation equivariance rule. However, both the temporal and heterogeneous challenges are not addressed. 

This work is thus motivated, and the key research difficulty is the dynamic evolution of communities embodied in heterogeneous graphs at different time steps.
Therefore, it is desired to learn effective feature representations which could be used for the discovery of ``varying'' communities, imposed from a series of graphs, in addition to ``static'' communities presented in current graph. Apparently, the desired feature representation should contain two parts. One part represents ``static'' features extracted from current graph. Another part representing ``dynamic'' features should be extracted from the interactions between graphs at different time steps. 
By doing so, it inevitably results in a large sparse feature representation space which requires a high computational cost. As a consequence, how to further reduce the high computational cost should be carefully considered during the model design process. 


To well address aforementioned research difficulties, we propose the heterogeneous temporal graph convolutional networks (HTGCN) for community detection task. In particular, we first define the calculation method to compute heterogeneous adjacency matrices, and then embed both spatial structural information and node features for each momentary graph separately. 
Then, we propose the \underline{Res}idual \underline{C}ompressed \underline{A}ggregation \underline{C}omponent (ResCAC). 
Inspired by \cite{Wang:2019:HAN}, this component first utilizes meta-paths to sample heterogeneously correlated nodes, potentially belonging to the same communities, from several consecutive graphs. This sampling step is analogous to edge conversion step in LGNN with the merit of preserving temporal community information.
After that, this component interacts node features and compresses these interactive feature representations. To simultaneously represent features of both time-invariant and evolutionary communities, we also aggregate features of current graph with the compressed temporal features, and the aggregation is achieved through a linear layer of residual connections \cite{he2015deep}. At last, a revised  community detection loss function is proposed.  


The major contributions of this paper can be summarized as follows:
\begin{itemize}
\item We propose the HTGCN model. To the best of our knowledge, this is among the first attempts to perform community detection on learnt feature representations of heterogeneous and temporal graph data. 
\item We utilize meta-path structure to sample nodes from two consecutive heterogeneous graphs, 
and then design a residual compressed aggregation component to build the feature representations for both ``static'' and ``dynamic'' features. 

\item We perform extensive experiments on two real-world datasets and the promising results have demonstrated that our HTGCN is superior to both benchmark and the state-of-the-art approaches.
\end{itemize}

\begin{figure}[t]
\centering
\includegraphics[height=1.3in, width=0.48\textwidth]{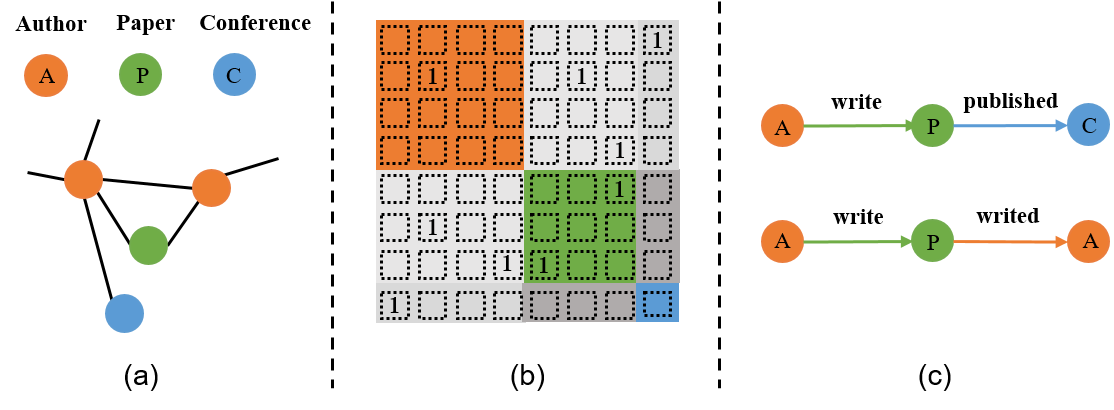}
\caption{An illustrating example using DBLP datasets. (a) A heterogeneous graph with different node types, e.g., author(A), paper(P) and conference(C). (b) The corresponding heterogeneous adjacency matrix. (c) Two meta-path examples: $A \to{P} \to{C}$ and $A\to{P}\to{A}$.}
\label{fig:example}
\vspace{-0.4cm}
\end{figure}

\section{Preliminaries and Problem Setup}
\subsection{Preliminaries}
\textbf{Heterogeneous Graph.} Let ${G}=(V,E)$ denote an undirected graph, where ${V}=\left\{v_{ij}\right\}$ and ${E} \subset {V} \times {V}$ denote node set and edge set, respectively. 
Let ${T}_{V}$ and ${T}_{E}$ respectively denote the set of node types and edge types, if $|{T}_{V}|+|{T}_{E}|>2$, ${G}$ is a heterogeneous graph. \\ 
\textbf{Adjacency Matrix of Heterogeneous Graph.} Similar to \cite{spectralclusteringHIN2015}, we define adjacency matrix of heterogeneous graph $G$ as $A_{HG} = \left\{A_{ij}|i \in [1, m],j \in[1, n]\right\}$, where $A_{ij}$ is an element entry, $m=|{T}_{V}|$ is the number of node types and $n=|{T}_{E}|$ is the number of edge types. Note that $A_{ij}$ denote an adjacency matrix of the generated homogeneous graph by fixing  node type $i$ and edge type $j$. 
Thus, the degree matrix of heterogeneous graph ${G}$ can be defined as $D_{HG}=diag(D_{ii})$, where $D_{ii}=\sum_{j} A_{ij}$ . That is, we sum up the number of edges of different types $j$ between two nodes of the same type $i$. \\
\textbf{Meta-Path.} Meta-path is originally proposed to capture relationships between heterogeneous nodes and edges. Generally, a meta-path $\delta$ could be defined as
\begin{eqnarray}
\nonumber \delta = a_1\stackrel{e_1}{\longrightarrow}a_2\stackrel{e_2}{\longrightarrow}\ldots\stackrel{e_{n-1}}{\longrightarrow}a_n,
\end{eqnarray}
where $a_1,a_2,...,a_n$ represent nodes of different types which are connected by edges $e_1,e_2,...,e_{n-1}$ of different types. For example, as shown in Figure \ref{fig:example}, a meta-path $A\to{P}\to{C}$ defined in DBLP dataset means that an author $A$ writes a paper $P$ published at conference $C$.

\subsection{Problem Setup}
To detect communities from graph, we denote $\mathbb{G}$ as a series of temporal graph data, and we have $\mathbb{G}=(\mathcal{G}^1,\mathcal{G}^2,...,\mathcal{G}^\mathcal{T})$ where $|\mathcal{T}|$ is the number of time steps. Let $\mathcal{G}^t = (V^t,A_{HG}^t,X^t,T,\mathcal{C}^{t})$ denote the heterogeneous graph at time step $t$, $V^t$ denote the node set of $\mathcal{G}^t$, $X^t \in \mathbb{R}^{N^t\times{D^t}}$ represent feature matrix of node set $V^t$ where $N^t$ represents the number of nodes of $\mathcal{G}^t$ and $D^t$ is the feature dimension of $X^t$, and $\mathcal{C}^t=\{1,..,C\}$ denote a set of community labels. The task of community detection in this paper is to label nodes in $\mathcal{G}^\mathcal{T}$ using the proposed HTGCN trained on a series of heterogeneous temporal graphs $\{\mathcal{G}^1,\mathcal{G}^2,...,\mathcal{G}^{\mathcal{T}-1}\}$ in a supervised manner by minimizing the objective function taking below general form, given as
\begin{equation}
Loss = \mathcal{L} (\mathcal{C}^t=\{1,..,C\}|X^t,\{\mathcal{G}^1,\mathcal{G}^2,...,\mathcal{G}^{\mathcal{T}-1}\})
\end{equation}

\section{The Proposed HTGCN}

\begin{figure}[t]
\setlength{\abovecaptionskip}{3pt}
\setlength{\belowcaptionskip}{0pt}
\centering
\includegraphics[height=1.8
in, width=0.48\textwidth]{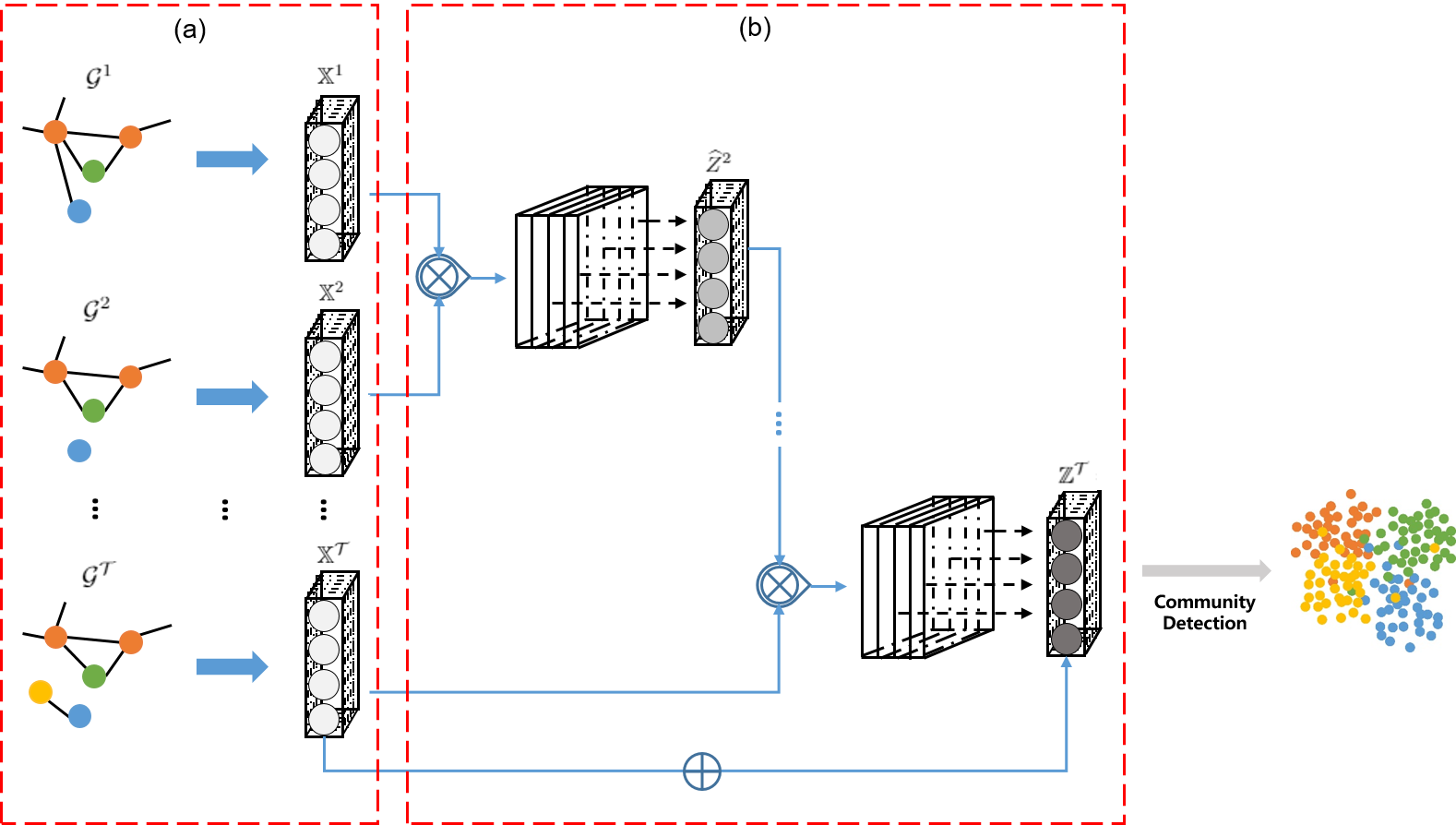}
\caption{The framework of the proposed HTGCN. It consists of two components: (a) Heterogeneous GCN component embeds neighbors' spatial information; (b) ResCAC component learns a compact set of ``dynamic'' features.}
\label{fig:framework}
\vspace{-0.4cm}
\end{figure}

As aforementioned, the proposed HTGCN is to detect communities out of a series of heterogeneous and temporal graph data. To cope with this issue, we first embed node features as well as their spatial structural information, globally calculated on the heterogeneous graph $\mathcal{G}^t$, 
into low-dimensional feature space. Then, a neural network component called \underline{Res}idual \underline{C}ompressed \underline{A}ggregation \underline{C}omponent (ResCAC) is proposed to aggregate the embedded features of consecutive temporal graphs. 
After acquiring these ``dynamic'' feature representations, we aggregate them with ``static'' features of $\mathcal{G}^\mathcal{T}$ to detect communities contained in $\mathcal{G}^\mathcal{T}$, and the corresponding framework of the proposed HTGCN is depicted in Figure \ref{fig:framework}.


\subsection{Heterogeneous GCN component}
To embed both spatial structural information and node features, we propose this heterogeneous GCN component which convolutes node features with the features of all its one-hop neighbors. According to \cite{kipf2017semi-supervised}, our convolutional operation could be defined as
\begin{eqnarray}
X_{i+1}=\sigma[\widehat{D}_{HG}^{-\frac{1}{2}}\left(A_{HG}+I\right)\widehat{D}_{HG}^{-\frac{1}{2}}X_{i}W_i],
\end{eqnarray}
where $\widehat{D}_{HG}=\sum_{j}(A_{ij}+I_{ij})$, $W_i$ is a weight matrix, $X_i$ is a feature matrix of the $i^{th}$ layer, and $\sigma$ is the ReLU activation function. The output of this component is denoted as $\mathbb{X}\in \mathbb{R}^{N\times{d}}$. 
$A_{HG}$ denotes a heterogeneous adjacency matrix which is generally used to represent neighborhood relationship among nodes of a specific node or edge type. 

\begin{figure*}[t]
\setlength{\abovecaptionskip}{6pt} 
\setlength{\belowcaptionskip}{0pt}
\centering
\includegraphics[height=1.8in, width=0.7\textwidth]{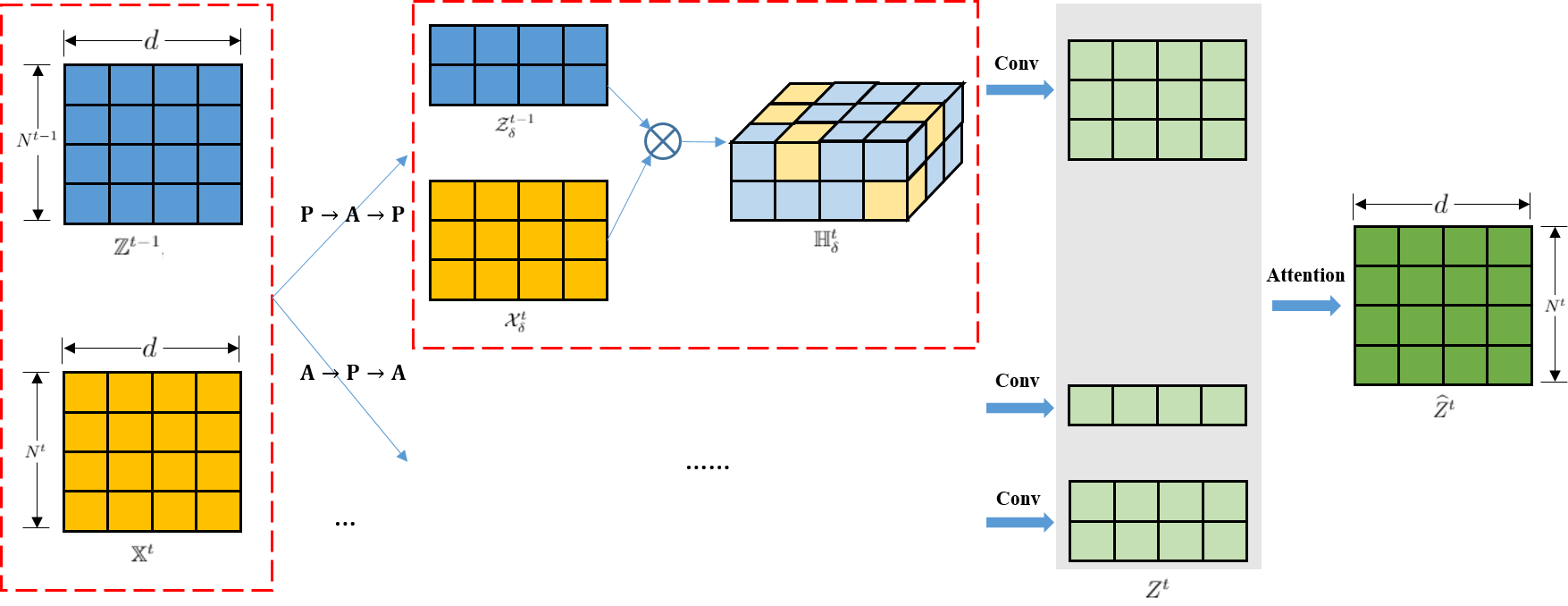}
\caption{Details of the proposed ResCAC component. First, we separately sample two feature representation matrices $\mathcal{Z}_{\delta}^{t-1}$ and $\mathcal{X}_{\delta}^t$ using the same meta-path ($P\to{A}\to{P}$, etc.). Second, we acquire a 3D tensor $H_{\theta}^i$ by Hadamard product and concatenate operation. Third, a one-dimensional convolution is applied to compress the tensor. Finally, an attention component is applied on $Z^t$ to differentiate weights of different features.\blue{
}
}
\label{fig:ResCan}
\vspace{-0.3cm}
\end{figure*}

\begin{algorithm}[tb]
\caption{The overall process of the HTGCN}
\label{alg:algorithm}
\textbf{Input}: A series of heterogeneous temporal graph

\setlength{\parindent}{3em} $\mathbb{G}=(\mathcal{G}^1,\mathcal{G}^2,...,\mathcal{G}^\mathcal{T})$,

\setlength{\parindent}{3em} Node feature \{$X^t, t\in[1,\mathcal{T}]$\},

Heterogeneous adjacency matrix \{$A_{HG}^t, t\in[1,\mathcal{T}]$\},

Meta-path \{$\delta^t, t\in[1,\mathcal{T}]$\}.\\
\textbf{Parameter}: Weight matrix \{$W_0,W_1$\},

\setlength{\parindent}{5em}  The number of meta-path $M$,

The number of nodes $N_t$ at time step $t$.\\
\textbf{Output}: The feature representations $\mathbb{Z}^t$.

\begin{algorithmic}[1] 
\STATE Heterogeneous GCN embedding component:
\FOR{$X^t \in \{X^1, X^2, ..., X^{\mathcal{T}}\}$}
\STATE
$\widehat{A}_{HG}^t \leftarrow{\widehat{D}_{HG}^{-\frac{1}{2}}\left(A_{HG}+I\right)\widehat{D}_{HG}^{-\frac{1}{2}}}$;\\
$\mathbb{X}^t \leftarrow{\widehat{A}_{HG}^t[{\rm ReLU}(\widehat{A}_{HG}^t X^t W_0)]W_1}$;
\ENDFOR
\STATE Residual Compressed Aggregation component:
\FOR{$\mathbb{X}^t \in \{\mathbb{X}^2, ..., \mathbb{X}^{\mathcal{T}}\}$}
\STATE $Z^t \leftarrow{\mathbb{X}^t}$
\FOR{$\delta_i \in \{\delta_1, \delta_2, ..., \delta_M\}$}
\STATE Sample feature representation matrix $\mathcal{Z}_{\delta_i}^{t-1}, \mathcal{X}_{\delta_i}^{t}$;\\
$\mathbb{H}_{\delta_i}^t \leftarrow{\mathop{\big\Vert}\limits_{k=1}^{d}\mathop{\big\Vert}\limits_{j=1}^{d}\mathcal{Z}_{\delta_i}^{t-1}(k,*)\circ{\mathcal{X}_{\delta_i}^{t}(j,*)}}$;\\
$Z^t \leftarrow{Z^t + {\rm Sigmoid}(\mathbb{H}_{\delta_i}^t\cdot{\widetilde{W}_k^t})}$;\\
\ENDFOR
\FOR{$i \in N_t$}
\STATE Calculate the weight coefficient $\widehat{a}_i^t$ in attention module;
\ENDFOR
\STATE $\widehat{A}^t \leftarrow{\mathop{\big\Vert}\limits_{i=1}^{N_t}\widehat{a}_i^t} $;\\
$\mathbb{Z}^t \leftarrow{\widehat{A}^t Z^t}$;
\ENDFOR
\STATE $\mathbb{Z}^\mathcal{T} \leftarrow{\mathbb{Z}^\mathcal{T} + \mathbb{X}^\mathcal{T}}$
\STATE \textbf{return} $\mathbb{Z}^\mathcal{T}$;
\end{algorithmic}
\end{algorithm}
\vspace{-0.3cm}

\subsection{Residual Compressed Aggregation Component}
After acquiring $\mathbb{X}$ generated by the heterogeneous GCN component, the proposed ResCAC component is essentially to learn feature representations which can well represent evolutionary patterns of the hidden communities. Note that the proposed ResCAC component is performed in a self-iterative manner, and it consists of three main operations, i,e. meta-paths, Hadamard product operation and a compression operation, detailed in the following paragraphs.  

First, we utilize meta-paths to sample heterogeneously correlated nodes from each momentary graph. Intuitively, although nodes might join or leave the heterogeneous graph at each time step, there still exist some time-invariant nodes. These retained nodes naturally become ``ties'' between graphs at different time steps. Therefore, we utilize meta-paths to sample heterogeneously correlated nodes. For instance, according to the meta-path $\delta$: $P\to{A}\to{P}$, we can screen out author nodes (type of `$A$'), which exist in $\mathcal{G}^{t-1}$ and $\mathcal{G}^t$. Then, we separately sample the embedded paper nodes (type of `$P$') into two feature matrices, i.e., $\mathcal{X}_{\delta}^{t}$ (for $P$ in $A\to P$ in $\mathcal{G}^t$) and $\mathcal{Z}_{\delta}^{t-1}$ (for $P$ in $P\to A$   in $\mathcal{G}^{t-1}$), for the same authors from two consecutive graphs, as illustrated in middle-upper red dotted rectangle in Figure \ref{fig:ResCan}. 

Second, we perform Hadamard product operation on two sub feature matrices of these nodes, respectively extracted from feature matrices $\mathbb{X}^{t}$ and $\mathbb{Z}^{t-1}$, to acquire a 3D tensor preserving the interactive features of ``varying'' communities. With $\mathcal{X}_{\delta}^{t}$ and $\mathcal{Z}_{\delta}^{t-1}$, the feature matrix after Hadamard product and concatenation is given as 
\begin{eqnarray}
\mathbb{H}_{\delta}^t = \mathop{\big\Vert}\limits_{i=1}^{d}\mathop{\big\Vert}\limits_{j=1}^{d}\mathcal{Z}_{\delta}^{t-1}(i,*)\circ{\mathcal{X}_{\delta}^{t}(j,*)},
\end{eqnarray}
where $2\leq{t}\leq{\mathcal{T}}$, $\mathbb{H}_{\delta}^t$ denotes a 3D tensor. $\mathcal{X}_{\delta}^{t} \in \mathbb{X}^t$ and $\mathcal{Z}_{\delta}^{t-1} \in \mathbb{Z}^t$ respectively denote feature representations of nodes generated by heterogeneous GCN component at time step $t$ and ResCAC at time step $t-1$, and $\mathop{\big\Vert}$ is the concatenation operator. Note that each 3D tensor is generated for each meta-path $\delta$, and thus the number of tensors is the same as the number of meta-paths.

Last, a compression operation is proposed to perform on $\mathbb{H}_{\delta}^t$, as illustrated in right-hand side grey rectangle in Figure \ref{fig:ResCan}.
To this end, the one-dimensional convolution operation is performed on each layer along the direction of feature dimension, and thus the 3D tensor is compressed into a 2D tensor $Z^t$, calculated as
\begin{eqnarray}
Z^t=\sum\limits_{i=1}^{M}\sum\limits_{k=1}^{N_{\delta_i}^t}\sigma(\mathbb{H}_{\delta_i}^t\cdot{\widetilde{W}_k^t}),
\end{eqnarray}
where $\widetilde{W}_k^t \in \mathbb{R}^{1\times(N_{\delta}^{t-1}\cdot{N_{\delta}^t)}}$ is the parameter matrix for the $k$-th layer (the $k$-th convolutional kernel) of the one-dimensional convolution. $N_{\delta_i}^t$ is the number of nodes at time step $t$ in meta-path $\delta_i$. $|M|$ is the number of meta-path, and $\sigma$ is sigmoid activation function. To differentiate the weight of different features representations, an attention component is naturally applied on $Z^t$, and the corresponding coefficient matrix is computed as 
\begin{eqnarray}
\widehat{A}^t = \frac{{\rm exp}\{\widehat{W}{\rm tanh}(\widehat{V}Z^t)\}}{\sum_{i=1}^{N^t}{\rm exp}\{\widehat{W}{\rm tanh}(\widehat{V}Z^t)\}},
\end{eqnarray}
where $\widehat{W} \in \mathbb{R}^{d_a}$ and $\widehat{V} \in \mathbb{R}^{d_a\times{d}}$ are parameter matrices. Consequently, the weighted output is now recalculated as 
\begin{eqnarray}
\widehat{Z}^t = \widehat{A}^t{Z^t} \in \mathbb{R}^{N^t\times{d}}.
\end{eqnarray}

To avoid high computation cost and the vanishing gradient issue, a natural choice is to adopt the ResNet \cite{he2015deep} structure. Therefore, we add up the original feature matrix to the model output $\widehat{Z}^\mathcal{T}$, and we have   
\begin{eqnarray}
\mathbb{Z}^\mathcal{T} = \widehat{Z}^\mathcal{T} + \mathbb{X}^\mathcal{T}.
\end{eqnarray}

\subsection{Community Detection}\label{tab: loss}
In this subsection, we define a loss function for community detection similar to \cite{chen2019supervised}. Without loss of generality, this paper considers the non-overlapping community detection problem. 
According to \cite{chen2019supervised}, node labels should satisfy the permutation equivariance property for community detection, and thus the loss function should be designed according to permutations of node labels. A softmax ($\cdot$) function is applied on the output feature representations $\mathbb{Z}^\mathcal{T}$. Let $\hat{c}_i$ denote the predicted community label of node $i$, $c_i$ denote the ground truth community label and $\mathcal{S}_{\mathcal{C}}$ denote the permutations, the loss function of the proposed model is given as 
\begin{eqnarray}\label{eq:loss1}
\mathcal{L} = \mathop{\rm INF}\limits_{\pi \in \mathcal{S}_{\mathcal{C}}} - \sum\limits_{i \in V}\pi(c_i){\rm log}(\hat{c_i}),
\end{eqnarray}
where $\pi$ is a set of permutations in $\mathcal{S}_\mathcal{C}$.
Assume that ${\Phi}:X\to{\mathcal{C}}$ to be the true mapping function between from $X$ to the ground truth labels $\mathcal{C}$. And ${\Omega}:X\to{\mathcal{\hat{C}}}$ is our mapping function from $X$ to the predicted labels $\mathcal{\hat{C}}$. By substituting these functions into Eq. \ref{eq:loss1}, we have 
\begin{eqnarray}\label{eq:loss2}
\mathcal{L} = \mathop{\rm INF}\limits_{\pi \in \mathcal{S}_{\mathcal{C}}} - \sum\limits_{i \in V}\Phi{\big({\pi(x_i)}\big)}{\rm log}\big[\Omega{\big(\pi(x_i)\big)}\big].
\end{eqnarray}
To optimize this equation is equivalent to optimize the second term.
Considering the fact that the number of communities in real-world graph data might be relatively large, it is essential if we can further reduce computational cost. A suggested method is to cluster label set $\mathcal{C}$ into ${\mathcal{\hat{C}}}$ sub groups, then we can approximately optimize the above objective function. The loss is back-propagated to update parameters of the proposed HTGCN. 

\begin{table}[t]
\setlength{\abovecaptionskip}{5pt}
\setlength{\belowcaptionskip}{0pt}
\centering
\begin{tabular}{lrrrr}
\toprule
\textbf{Dataset} & \textbf{Nodes} & \textbf{Edges} & \textbf{Features} & \textbf{Communities} \\
\midrule
DBLP       & 20919  & 117074  & 174  & 3 \\
IMDB       & 10114  & 55924   & 1213 & 5 \\
\bottomrule
\end{tabular}
\caption{The statistics of DBLP and IMDB datasets}
\label{tab:dataset}
\end{table}
\begin{table}[t]
\setlength{\abovecaptionskip}{5pt}
\setlength{\belowcaptionskip}{0pt}
\centering
\begin{tabular}{lll}  
\toprule
\textbf{Dataset}  & \textbf{Relations} & \textbf{Meta-paths} \\
\midrule
DBLP        & paper-author  & $P\to{A}\to{P}$      \\
            & paper-conference  & $P\to{C}\to{P}$   \\
            & author-conference  & $A\to{C}\to{A}$    \\
IMDB        & movie-actor        & $M\to{A}\to{M}$  \\
            & movie-director     & $M\to{D}\to{M}$  \\
            & actor-director     & $A\to{D}\to{A}$  \\
            & director-actor     & $D\to{A}\to{D}$  \\
\bottomrule
\end{tabular}
\caption{The relations and meta-paths defined in each dataset.}
\label{tab:metapath}
\vspace{-0.3cm}
\end{table}

\begin{table*}[tbp]
\setlength{\abovecaptionskip}{5pt} 
\setlength{\belowcaptionskip}{0pt}
\centering
\resizebox{0.8\textwidth}{29mm}{
\renewcommand{\arraystretch}{1.5}
\begin{tabular}{lrrrrrrrrrrrr}
\toprule[2pt]
\multirow{2}*{\textbf{Methods}}&
\multicolumn{6}{c}{\textbf{DBLP}} & \multicolumn{6}{c}{\textbf{IMDB}}\\
\cmidrule(lrlrlr){2-7}\cmidrule(lrlrlr){8-13}
&ACC & NMI & Modularity & ARI & Macro-F1 & Micro-F1 & ACC & NMI & Modularity & ARI & Macro-F1 & Micro-F1\\
\midrule
GCN & 93.00 & 80.90 & 61.13 & 86.20 & 95.31 & 95.33 & 56.11 & 44.56 & 33.79 & 21.31 & 60.82 & 64.89 \\
GAT & 92.80 & 80.15 & 60.11 & 82.13 & 93.93 & 93.87 & 50.14 & 41.45 & 30.27 & 23.60 & 60.95 & 60.11 \\
GNN & 92.20 & 79.59 & 60.33 & 85.48 & 94.50 & 94.80 & 60.31 & 48.73 & 30.27 & 25.94 & 71.09 & 65.44 \\
\hline
LGNN & 91.40 & 76.73 & 61.09 & 83.37 & 94.22 & 94.27 & 58.47 & 29.46 & 26.02 & \textbf{28.33} & 51.90 & 60.78 \\
HAN & 94.13 & 51.84 &	58.72 &	42.51 &	94.96 &	95.07 &	61.48 &	50.56 &	40.29 &	24.10 &	70.93 &	65.02 \\
STAR & 84.51 &	57.08 &	60.50 &	58.56 &	84.64 &	84.64 &	59.00 &	34.43 &	42.53 &	23.29 &	53.05 &	53.05  \\
\hline
HTGCN-3T &\textbf{ 98.93} &	\textbf{94.32}&61.12&	\textbf{96.83} &\textbf{98.91}&	\textbf{98.93}&	\textbf{66.00} &\textbf{51.17}&	43.38 &27.20 &	\textbf{74.33} &\textbf{66.00}\\
HTGCN-5T & 98.13 &	90.78 &	61.51& 	94.36 &	98.13 &	98.13 &	64.67 &	48.77 &	\textbf{44.98}& 	26.17 &	71.12 &	64.67  \\
HTGCN-7T & 97.87& 	89.88 &	\textbf{61.58} &	93.55& 	97.87 &	97.87 &	65.33 &	50.84 &	44.79 &	26.69 &	73.19 &	65.33  \\
\bottomrule[2pt]
\end{tabular}}
\caption{Community detection results for all compared methods (\%). The proposed HTGCN-3T, HTGCN-5T and HTGCN-7T respectively denotes how many heterogeneous and temporal consecutive graphs to be convoluted.} 
\label{tab:fullresult}
\vspace{-0.3cm}
\end{table*}

\section{Experiments}\label{sec:exp}
\subsection{Datasets}
To evaluate the performance of the proposed HTGCN, we implement our approach as well as other compared methods on two real-world datasets, i.e., DBLP\footnote{\url{https://dblp.uni-trier.de/}} and IMDB\footnote{\url{https://www.imdb.com/}} dataset, 
and statistics of these two datasets are reported in Table \ref{tab:dataset} and \ref{tab:metapath}, respectively. A number of widely adopted evaluation criteria are chosen in the experiments including \textit{Accuracy}, \textit{NMI}, \textit{Modularity}, \textit{ARI}, \textit{Macro-F1} and \textit{Micro-F1}.  

\begin{itemize}[leftmargin=*]
\item \textbf{DBLP} is a monthly updated citation network dataset. To construct the heteregeneous graph, we extract three kinds of node types from DBLP, i.e., paper(P), author(A) and conference(C). 
Three meta-paths are defined as $P\to{A}\to{P}$, $A\to{C}\to{A}$ and $P\to{C}\to{P}$. There are three communities to detect which are \textit{Information Retrieval}, \textit{Data Base} and \textit{Machine Learning}. 

\item \textbf{IMDB} 
is one of the most widely adopted datasets for heterogeneous graph analysis task. It consists of information about ``director'', ``actors'' and ``Movie Release Date''. In our experiments, we construct three types of nodes, i.e., movie(M), director(D) and actor(A), and four meta-paths, i.e., $M\to{A}\to{M}$, $M\to{D}\to{M}$, $A\to{D}\to{A}$ and $D\to{A}\to{D}$ to sample data. There are five movie communities to detect which are \textit{Action}, \textit{Adventure}, \textit{Comedy}, \textit{Crime} and \textit{Drama}.
\end{itemize}


\subsection{Baseline Methods}
For performance comparison, following baseline methods as well as the state-of-the-art approaches are implemented.

\begin{itemize}[leftmargin=*]
\item GCN \cite{kipf2017semi-supervised} is considered as a benchmark graph neural network approach, originally proposed for 
semi-supervised classification on homogeneous graph. 


\item GAT \cite{velikovi2018GAT} is proposed for 
modeling heterogeneous graph by applying a hidden self-attention layer to assign different weight to different node features.

\item GNN and LGNN proposed in \cite{chen2019supervised} are graph neural network based state-of-the-art community detection approaches. 
LGNN first converts community detection problem to node classification one, and this is the most related approach to ours.


\item HAN \cite{Wang:2019:HAN} is also proposed to model heterogeneous graph by discovering both node-level and semantic-level information via designed attention mechanisms. In the experiments, we construct corresponding meta-paths for each dataset according to the original HAN.  


\item STAR \cite{xu2019star} is a temporal approach to  learn feature representations of temporal attributed graph by a designed grated recurrent unit (GRU) network. 

\end{itemize}

\subsection{Experimental Settings}

Note that these compared approaches cannot be directly applied to model heterogeneous and temporal graph data. For these approaches, we merge all momentary graphs to form a global one for fair comparison and we then customize each approach by using our calculated heterogeneous adjacency matrix in all experiments. Apparently, this is a non-trivial task. 
To evaluate how the graphs at past time steps affect community detection task, we respectively convolute past three, five and seven graphs in an iterative manner to generate three versions of HTGCN denoted as HTGCN-3T, HTGCN-5T and HTGCN-7T. The heterogeneous GCN has two layers and $d$ is set to the number of underlying communities. All approaches are implemented using PyTorch and optimized by Adam \cite{kingma2014adam} with a learning rate of 0.001. 



\subsection{Results on Community Detection}

We evaluate all compared methods as well as our HTGCN-3T, HTGCN-5T and HTGCN-7T for community detection task and report the corresponding results on DBLP and IMDB datasets in Table \ref{tab:fullresult}. 
It is noticed that DBLP results of all approaches are  significantly better than IMDB results. The possible reasons might be as follows. First, \textit{authors} in DBLP data usually belongs to a specific research area, whereas \textit{actors} or \textit{directors} may belong to different communities. 
Second, it is worth noting that the number of node features in IMDB data is nearly 7 times that of DBLP data. Thus, such multi-class nature and high-dimensional features make it more difficult to detect communities from IMDB data.

\textbf{DBLP results.} We observe that HTGCNs significantly outperform all other approaches w.r.t. all evaluation criteria. Particularly, the \textit{NMI} score is improved by up to 17\% over the best model, i.e., GCN. Furthermore, the \textit{ACC} and \textit{Macro-F1} scores of HTGCNs indicate that the proposed approach could perfectly detect community label for each node. Compared with spatial information embedding based approaches like GCN, HTGCNs embed both spatial and temporal features which could explain their superior performance. We also notice that HAN achieves the second best score on ACC criterion. 
From DBLP results, we can conclude that it is crucial to simultaneously represent both heterogeneous and temporal features under a unified framework. 


\textbf{IMDB results.} 
Similar observations could be seen from IMDB results. However, LGNN achieves the best \textit{ARI} score while the rest criteria are still far from satisfying. One possible reason is that LGNN employs a so-called ``non backtracking'' operator which can well preserve the spatial node information across multiple steps, whereas HTGCNs cannot extract deeper spatial information which is restricted by the nature of meta-path. Moreover, we observe that longer time intervals could not further enhance model performance, as seen in the results of HTGCN-5T and HTGCN-7T. Although they could better preserve long-term temporal features, ``varying'' communities may not necessarily exist in current graph.


\begin{figure}[t]
\setlength{\abovecaptionskip}{5pt}
\setlength{\belowcaptionskip}{0pt}
\centering
\subfigure[Accuracy] {\includegraphics[height=1.3in,width=1.58in]{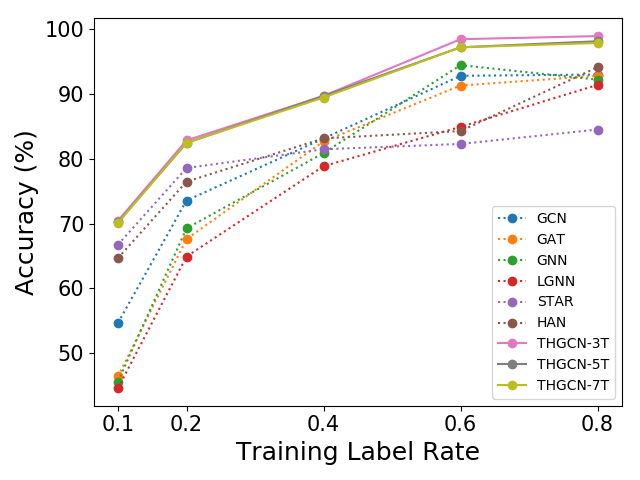}\label{result1}}
\subfigure[NMI] {\includegraphics[height=1.3in,width=1.58in]{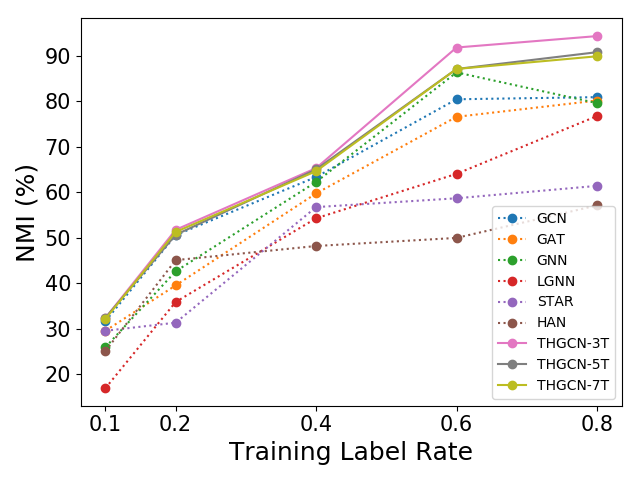}\label{result2}}
\caption{Effect of training label rates on DBLP dataset.}
\label{fig:labelRate}
\vspace{-0.3cm}
\end{figure}

\subsection{Effect of training label rates}
To investigate the effect of training label rates, we choose \textit{ACC} and \textit{NMI} criteria to evaluate all approaches on DBLP dataset. In this experiment, we respectively choose training label ratio to 10\%, 20\%, 40\%, 60\% and 80\% and report the corresponding results in Figure \ref{fig:labelRate}. From this figure, HTGCN-3T is constantly superior to the rest compared methods and the HTGCN-5T is the second best one. This further verifies the efficacy of the proposed approach. 


\subsection{Visualization Results}
Due to page limitation, we only visualize community detection results of the original GCN, HAN, LGNN and HTGCN, as plotted in Figure \ref{fig:visul}. 
Apparently, the HTGCN achieves the best visualization results on both datasets especially on IMDB. From the visualization results on IMDB, it is noticed that the five communities discovered by HTGCN could be well separated and spread over the plane, whilst the ``red'' community in Figure \ref{fig:visul} (a) is still mixed with the rest communities. Meanwhile, the ``green'', ``yellow'' and ''purple'' communities in LGNN overlap with each other, and thus is hard to separate. HAN can well separate ``red'' community but cannot spread the ``blue'' and ``orange'' communities. These two communities might be ignored in practical applications. We observe that, for visualization results on DBLP, all approaches could well separate the discovered communities. However, both GCN and HTGCN could spread these communities over 2D plane, and HTGCN is slightly better than GCN. This observation further verify the effectiveness of the proposed HTGCN. 

\section{Related Work}
Conventionally, community detection algorithms are usually topological structure based ones \cite{girvan2002community,raghavan2007near,rosvall2008maps}. 
Most recently, several state-of-the-art approaches have been proposed and the most related approaches are briefly reviewed as follows. 
\cite{he2019LDA-MRF} combines LDA and MRF under a unified framework to detect communities. 
Line Graph Neural Networks (LGNN) \cite{chen2019supervised} designs a polynomial function to calculate adjacency matrices for line graphs using the proposed non-backtracking operators for community detection. LGNN only fits for homogeneous graph data, whereas our proposed HTGCN well suits for heterogeneous graphs. HAN \cite{Wang:2019:HAN} models heterogeneous graph data by designing a semantic-level and a node-level attention component to learn the weight of neighboring nodes extracted using different meta-paths. HetGNN \cite{zhang2019Het-GNN} first employs random walk to sample heterogeneous neighboring nodes and then designs a graph neural network to aggregate features of these neighboring nodes. However, it does not consider the situation that nodes may vary with time which limits its potential applicability.  
With a focus on temporal data, STAR \cite{xu2019star} proposes a RNN model with one spatial attention component to embed features of important neighboring nodes, and another temporal attention component to filter out more important momentary graphs. Note that it cannot model heterogeneous graph data. 
Different from these related works, the proposed HTGCN embeds both spatial information and node attributes for a series of heterogeneous and temporal graphs. 
To the best of our knowledge, this is among the first attempts to detect community from heterogeneous and temporal graph data.






\begin{figure}[t]
\setlength{\abovecaptionskip}{5pt}
\setlength{\belowcaptionskip}{0pt}
\centering
\subfigure[GCN] {\includegraphics[height=0.86in,width=0.82in]{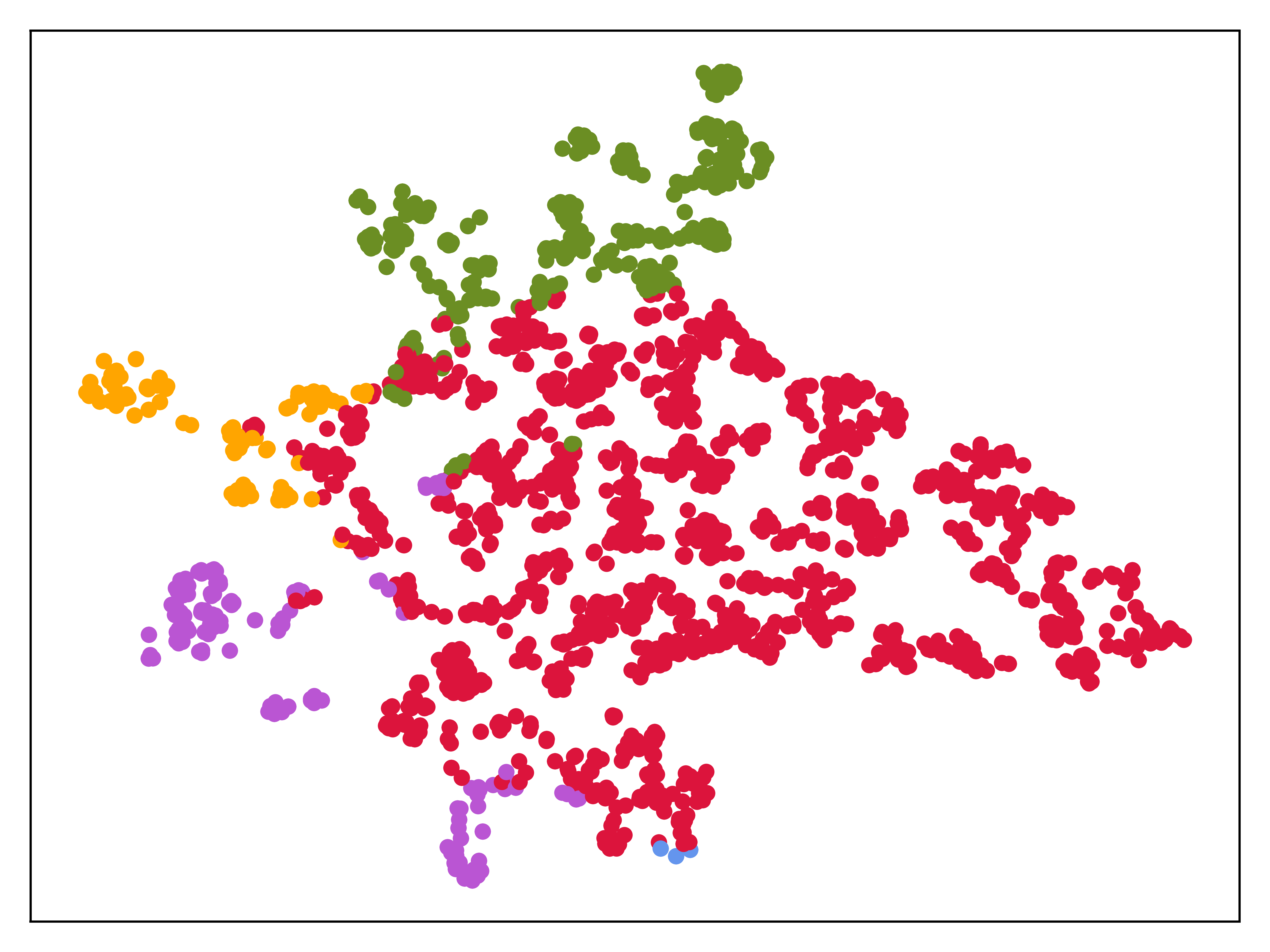}\label{fig:GCN-IMDB}}
\subfigure[LGNN] {\includegraphics[height=0.86in,width=0.82in]{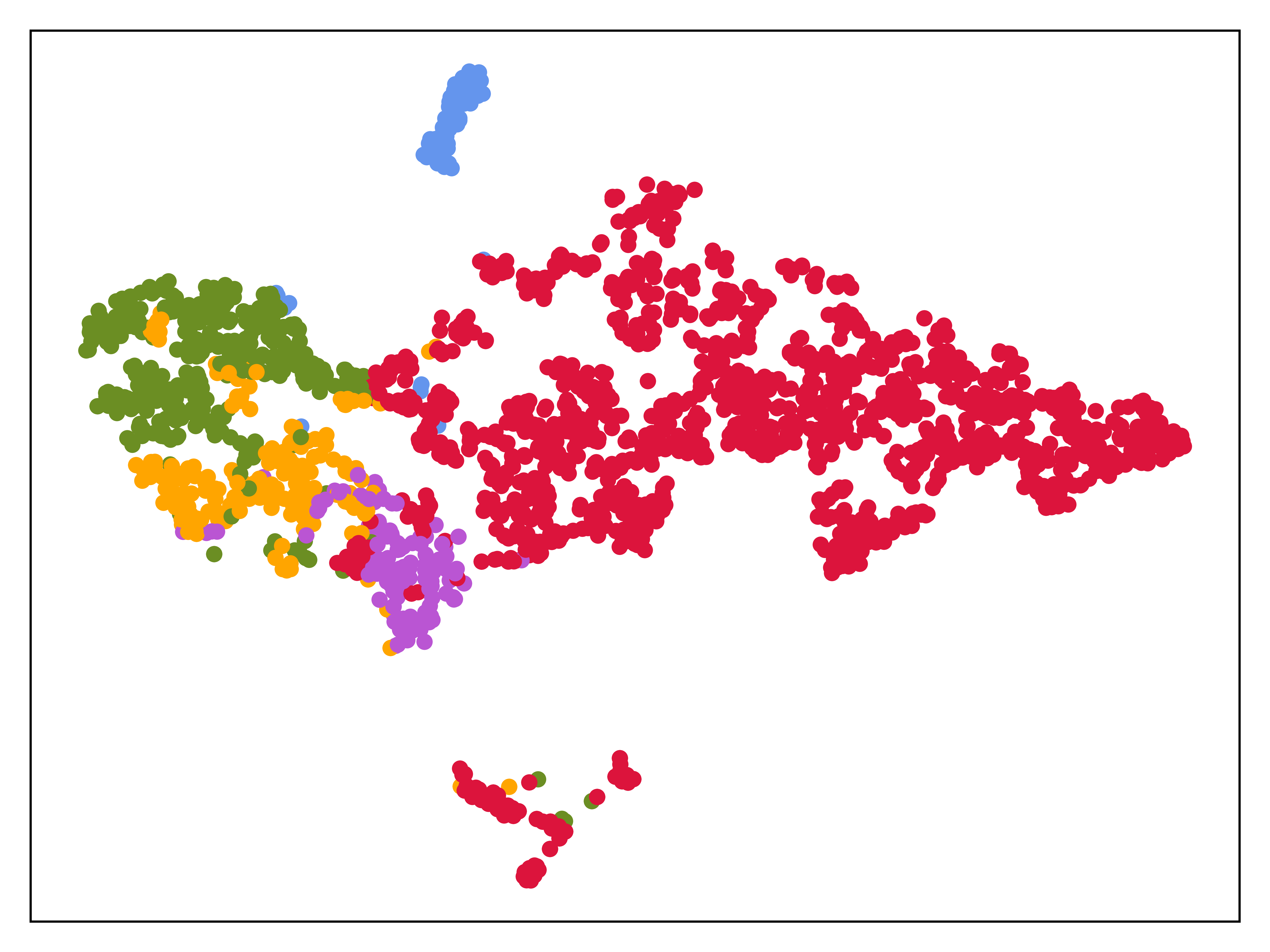}\label{fig:LGNN-IMDB}}
\subfigure[HAN] {\includegraphics[height=0.86in,width=0.82in]{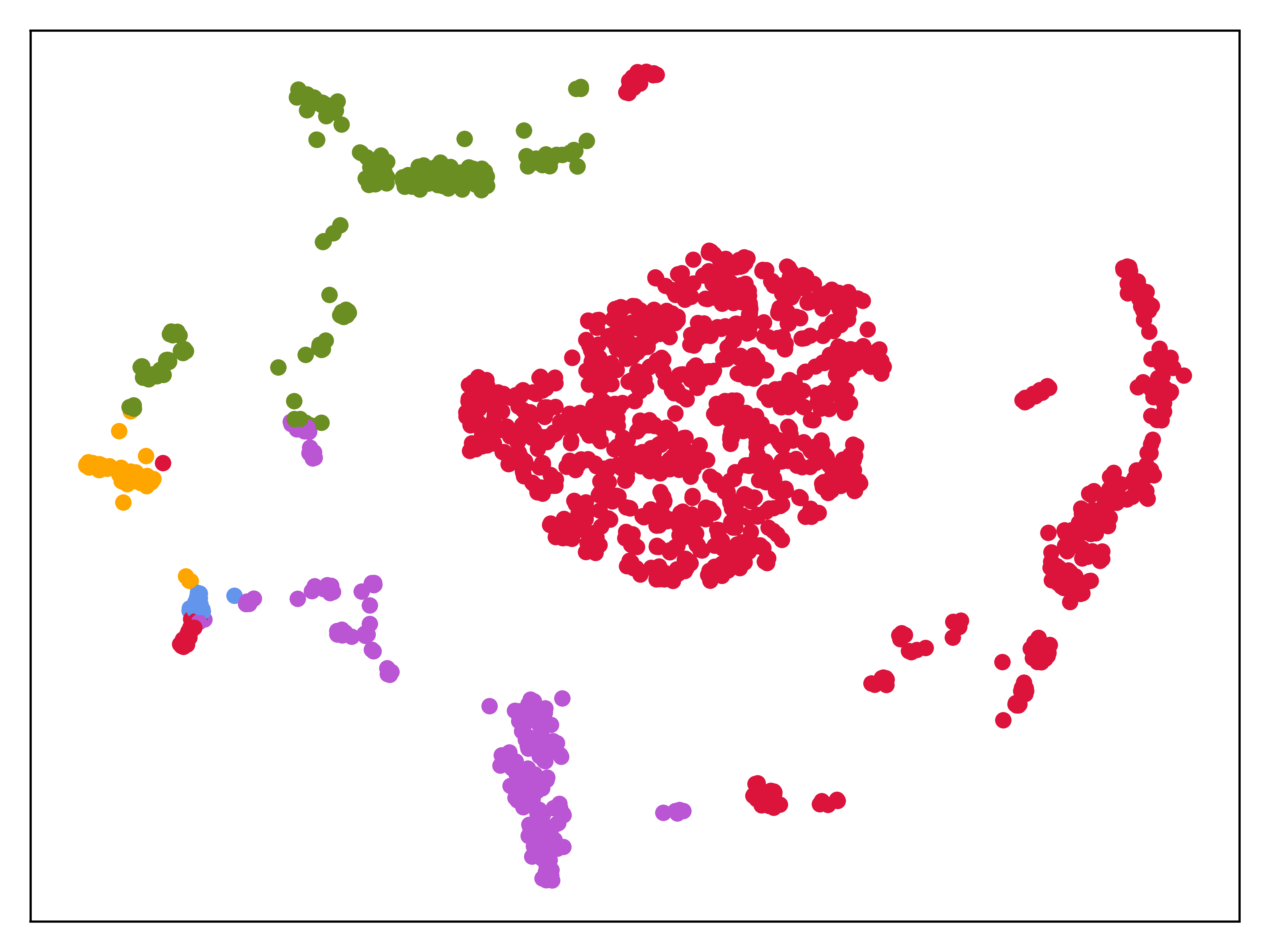}\label{fig:HAN-IMDB}}
\subfigure[HTGCN]
{\includegraphics[height=0.86in,width=0.82in]{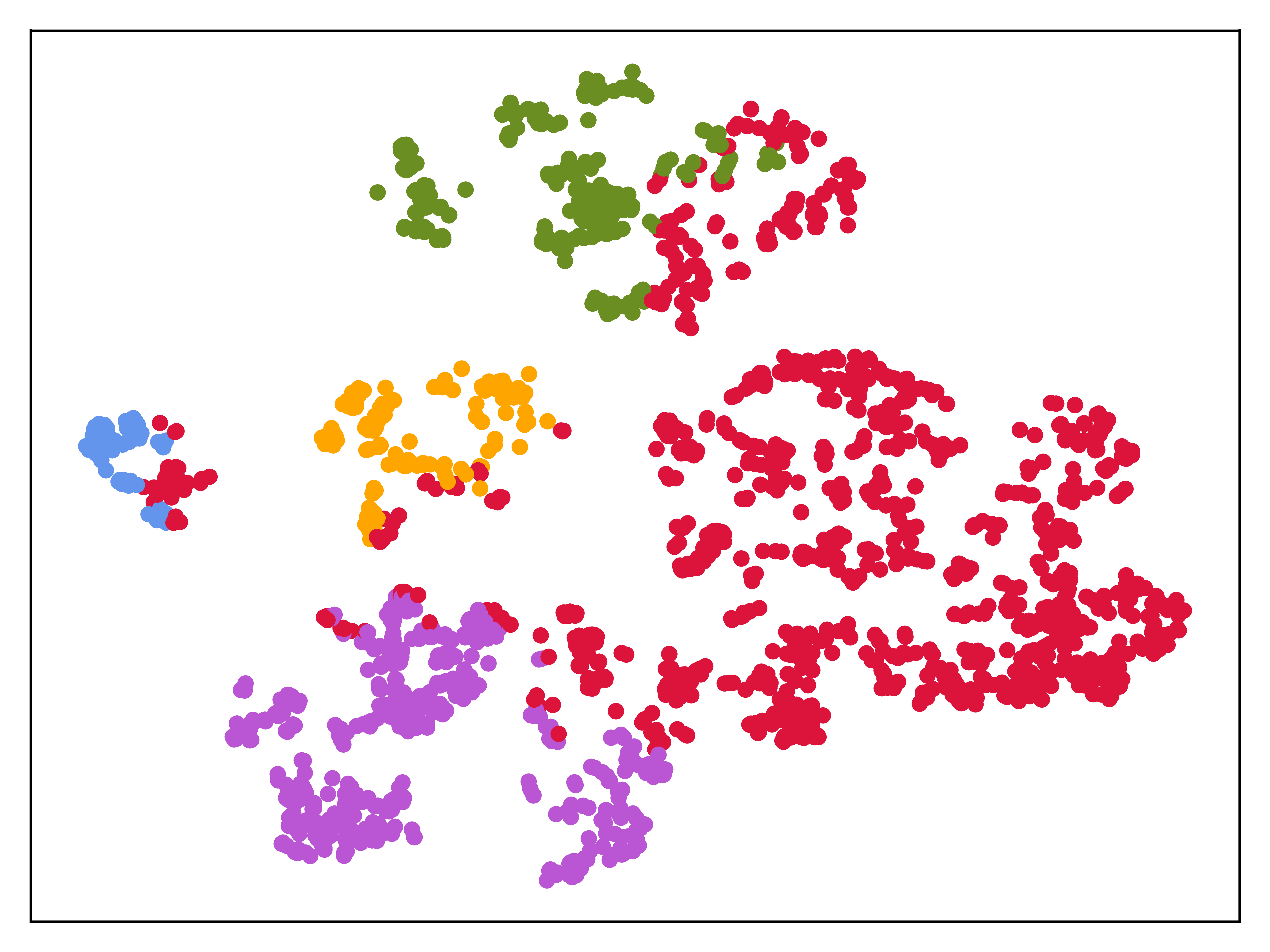}\label{fig:HTGCN-IMDB}}

\caption{Visulaization results on IMDB dataset.}
\label{fig:visul}
\vspace{-0.4cm}
\end{figure}

\begin{figure}[t]
\setlength{\abovecaptionskip}{5pt}
\setlength{\belowcaptionskip}{0pt}
\centering
\subfigure[GCN] {\includegraphics[height=0.86in,width=0.82in]{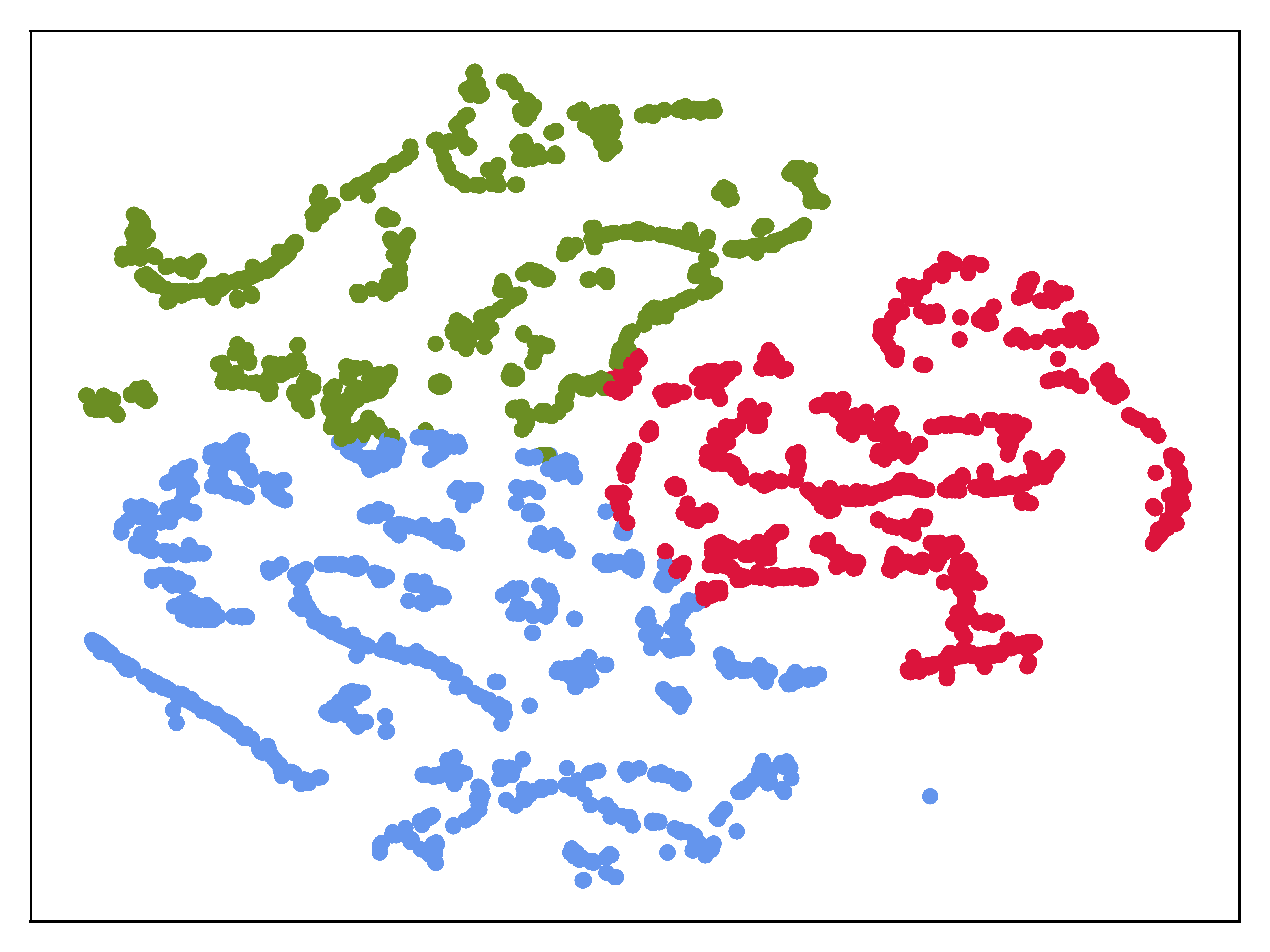}\label{fig:GCN-DBLP}}
\subfigure[LGNN] {\includegraphics[height=0.86in,width=0.82in]{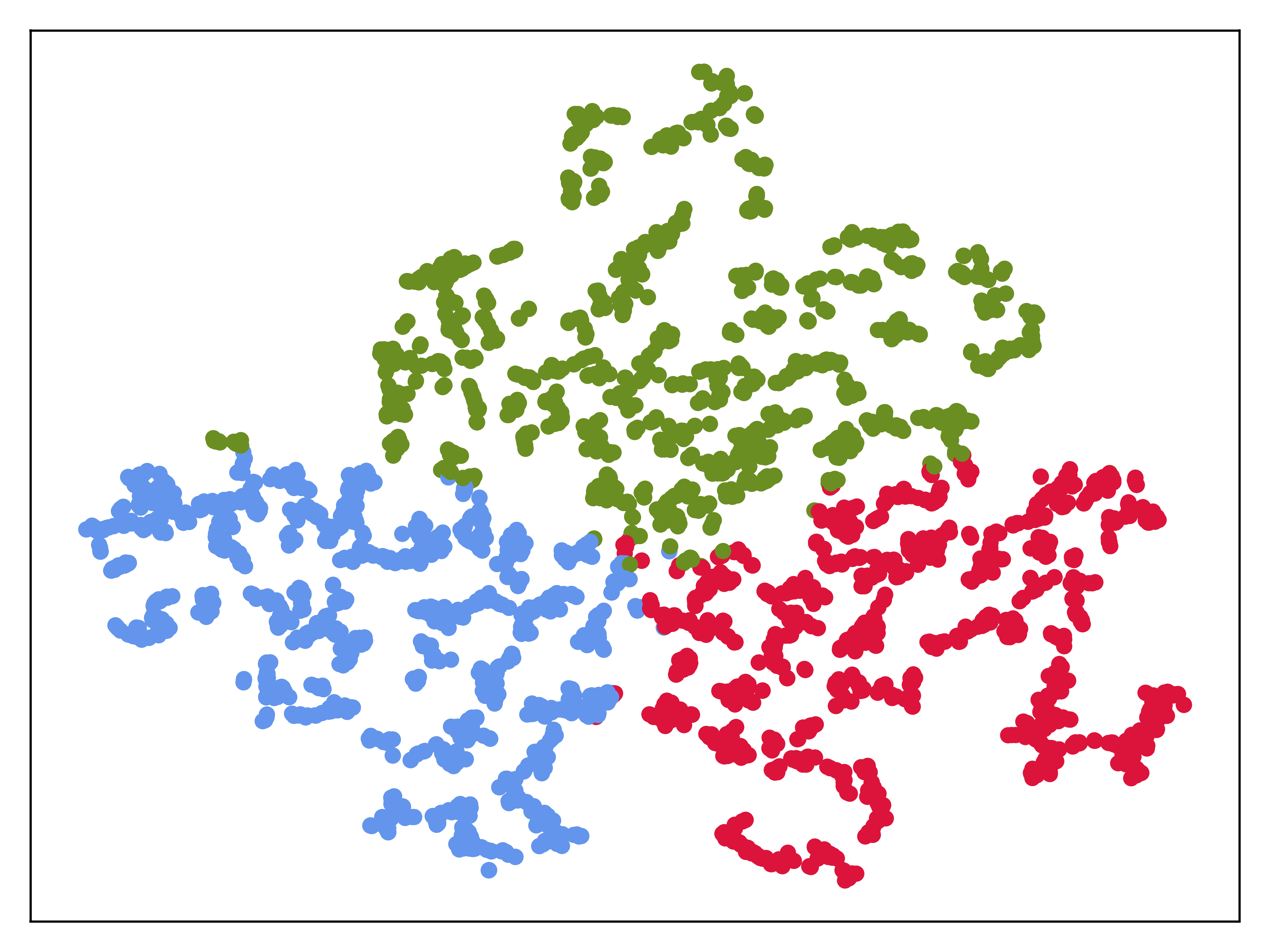}\label{fig:LGNN-DBLP}}
\subfigure[HAN] {\includegraphics[height=0.86in,width=0.82in]{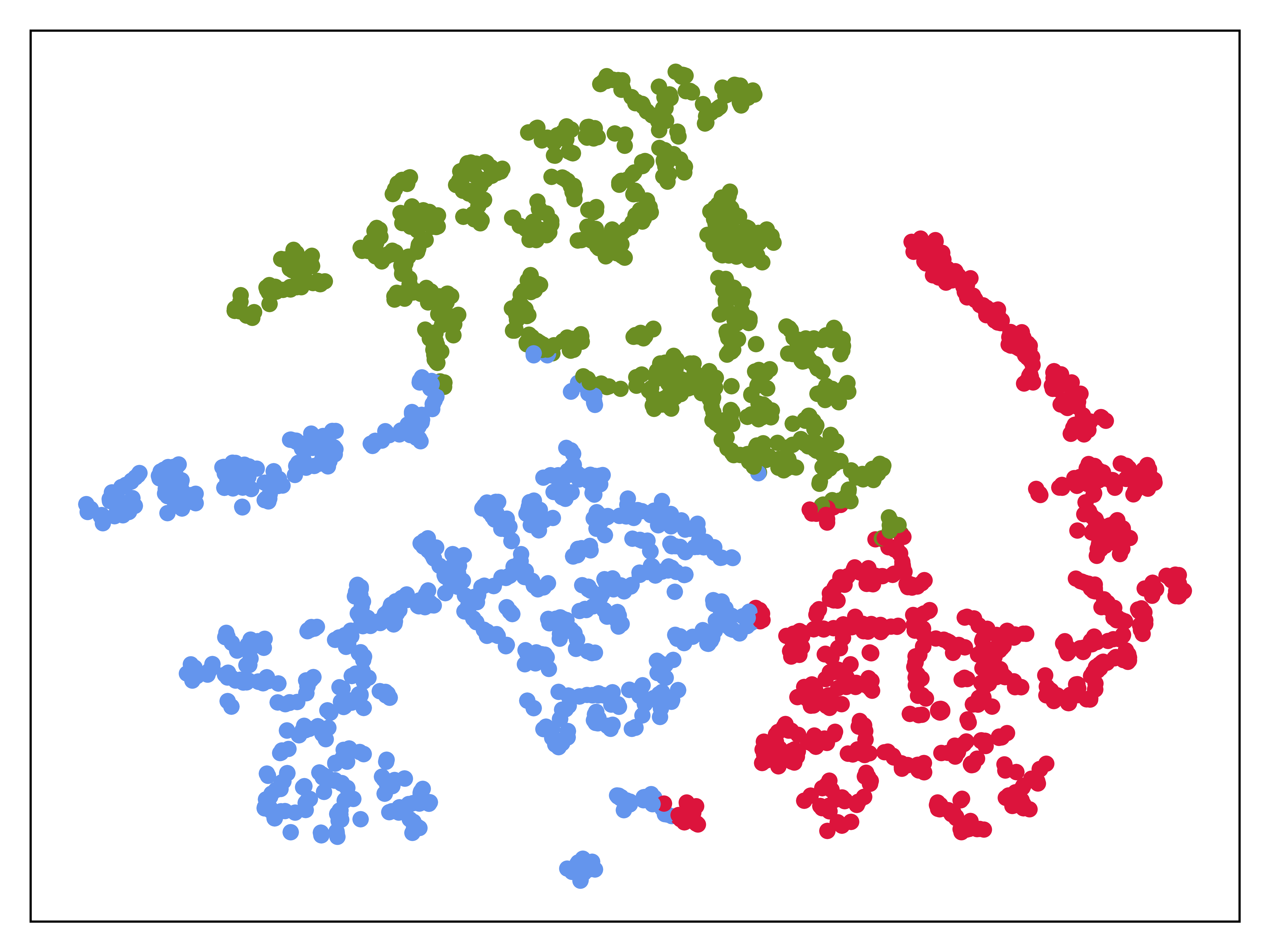}\label{fig:HAN-DBLP}}
\subfigure[HTGCN]
{\includegraphics[height=0.86in,width=0.82in]{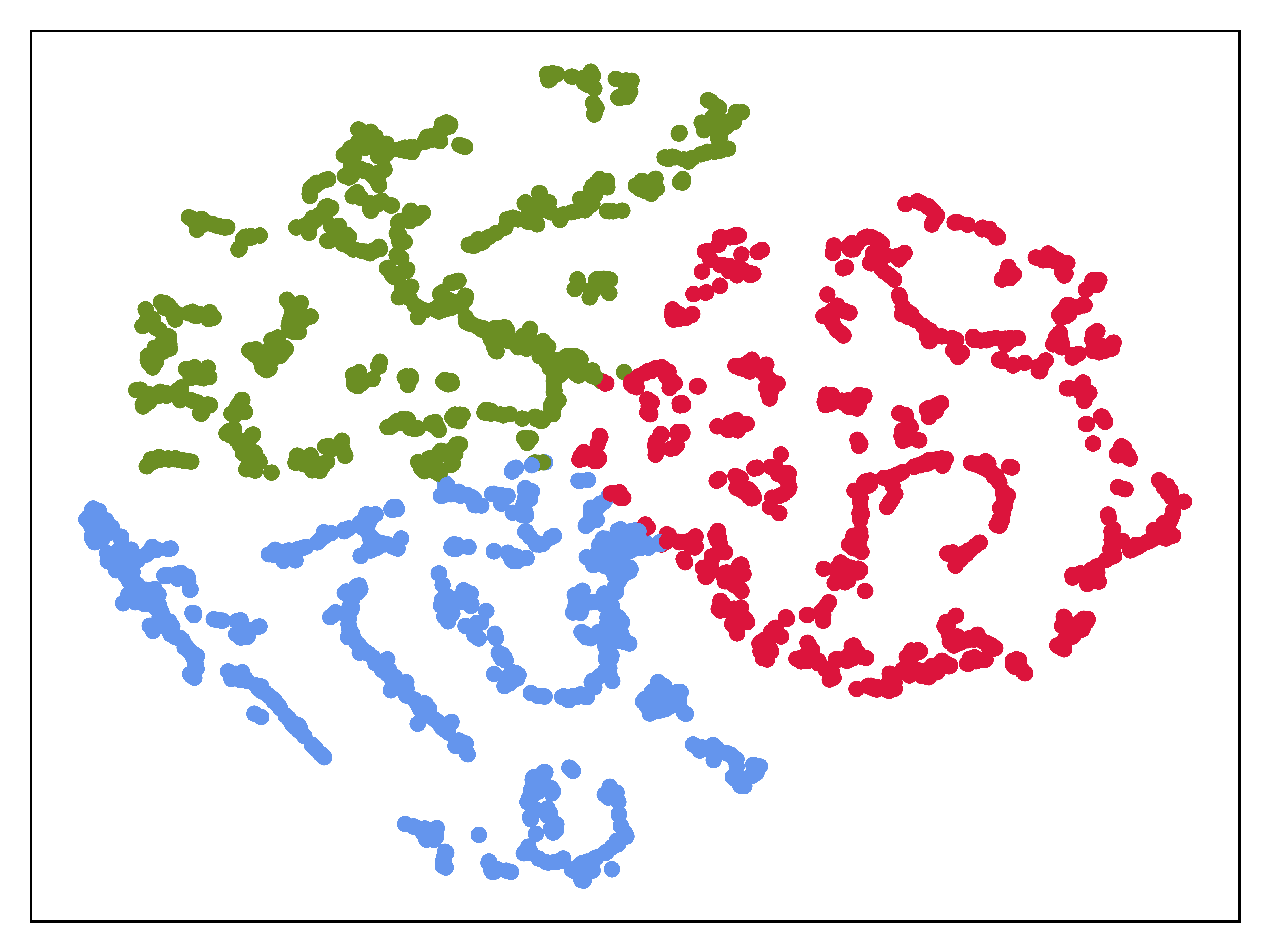}\label{fig:HTGCN-DBLP}}
\caption{Visulaization results on DBLP dataset.}
\label{fig:visul}
\vspace{-0.3cm}
\end{figure}

\section{Conclusion}
In this paper, we propose a novel heterogeneous-temporal graph convolutional networks (HTGCN) for community detection task. Particularly, a heterogeneous GCN component and a ResCAC component are proposed to learn feature representations for both ``static'' and ``dynamic'' features. Both community detection results and visualization results on two real-world datasets demonstrate that the HTGCN achieves the superior performance over the state-of-the-art approaches. 

~~\newpage{}
\bibliographystyle{named}
\bibliography{ijcai20}

\end{document}